\pdfoutput=1

\documentclass[letterpaper, 10 pt, conference]{ieeeconf}

\IEEEoverridecommandlockouts                              
\usepackage{graphicx} 

\usepackage[bookmarks=true]{hyperref}

\usepackage{tabularx}
\usepackage{booktabs}
\usepackage{multirow,array}

\title{\LARGE \bf
The Aqualoc Dataset: Towards Real-Time Underwater Localization from a Visual-Inertial-Pressure Acquisition System
}

\author{Maxime Ferrera$^{1,2}$, Julien Moras$^{1}$, Pauline Trouv\'e-Peloux$^{1}$, Vincent Creuze$^{2}$, Denis D\'egez$^{2}$
\thanks{This work was partially supported by the CNRS (Mission pour l'interdisciplinarit\'e - Instrumentation aux limites 2018 - Aqualoc project) and R\'egion Occitanie (ARPE Pilotplus project).}
\thanks{$^{1}$M. Ferrera, J. Moras and P. Trouv\'e-Peloux are with DTIS, ONERA, Universit\'e Paris Saclay F-91123 Palaiseau - France.}%
\thanks{$^{2}$M. Ferrera and V. Creuze are with LIRMM, Universit\'e de Montpellier, CNRS, Montpellier, France.}%
\thanks{$^{3}$D. D\'egez is with the DRASSM, Ministry of Culture, Marseilles, France}%
}

\begin{document}

\maketitle
\thispagestyle{empty}
\pagestyle{empty}




\section{INTRODUCTION}

Robots localization is a challenging task which is essential for any autonomous robot or remotely operated one.  In outdoor, GNSS signals can be used to get a quite accurate estimate of the position.  However, in GNSS denied environment, such as indoor or underwater, localization needs to be estimated from other sensors.
In aerial and terrestrial robotics, monocular VO (\textit{Visual Odometry}), VSLAM (\textit{Visual Simultaneous Localization And Mapping}) and, more recently, VI-SLAM (\textit{Visual-Inertial SLAM}) have shown great results \cite{ORB-SLAM,SVO-2,Leutenegger-OKVIS_ijrr,Vins-Mono}.  Filtered SLAM based on Extended Kalman Filters (EKF) have been set aside to the profit of Bundle Adjustment (BA) based SLAM \cite{WhyFilter}.  These SLAM methods, hugely relying on nonlinear optimization, have been successfully used to estimate localization from low-cost sensors such as cameras and MEMS-IMU (Micro-Electro-Mechanical System - Inertial Measurement Unit).  The availability of many public terrestrial or aerial datasets have helped a lot in the quick development of these methods.  For example, KITTI \cite{Kitti}, EuRoC \cite{Euroc} and Mono-VI TUM \cite{tumvidataset} are famous datasets dedicated to the study of VSLAM or VI-SLAM algorithms.

\noindent In this paper, we focus on vision-based SLAM methods in underwater environments.  The use of visual information is challenging underwater as the medium creates many visual degradations such as turbidity, back-scattering and light absorption.  These difficulties led most works to turn to sonar based SLAM systems \cite{Ekf4AuvSonar,slamInManMadeEnv}.  Nevertheless, some works have also demonstrated the potential of cameras for underwater localization \cite{stereo_graph_slam,CreuzeMonoVO}.  However, most of these SLAM methods rely on the integration of expensive navigational sensors (Doppler Velocity Logs, Fibber Optic Gyroscopes or high-end IMUs) to provide accurate enough localization information, using only sonars or cameras to bound the drift \cite{EusticeVAN, auto_mapping_ridao}.

\begin{figure}
\centering
\includegraphics[width=0.95\linewidth]{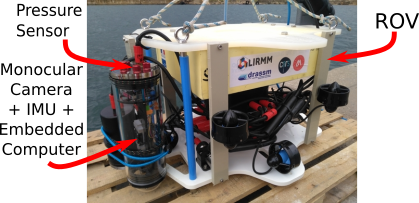}
\caption{The Remotely Operated Vehicle (ROV) and acquisition system used to record the dataset.}
\label{fig:rov_and_caisson}
\vspace{-5mm}
\end{figure}

\noindent We believe that underwater localization from low-cost sensors processed through BA based SLAM could open the way to new localization techniques.  However, the lack of underwater datasets is a limitation to the development of such algorithms.  The only underwater dataset with visual information allowing the use of VSLAM methods we are aware of is \cite{UWsimData} but it consists only of simulated images.  Hence, in order to open the way to deeper studies of these SLAM techniques, we propose a new underwater dataset that we make publicly available online\footnote{http://www.lirmm.fr/aqualoc/}.  This dataset has been recorded in an harbor and provides several sequences with synchronized measurements from a monocular camera, a MEMS-IMU and a pressure sensor.  To the best of our knowledge, this  is the first underwater dataset dedicated to the study of underwater localization methods from low-cost sensors.

\noindent The rest of this paper is organized as follow.  First, we present the payload designed for this acquisition.  Then, we give details about the recorded dataset.  Finally, we present results of an evaluation of state-of-the-art open-source monocular VO and VSLAM which can be used as a benchmark on this dataset and highlights the potential of such vision based localization methods.

\section{Acquisition system}
\label{sec:acq_sys}

\begin{figure}
\centering
\includegraphics[width=0.65\linewidth]{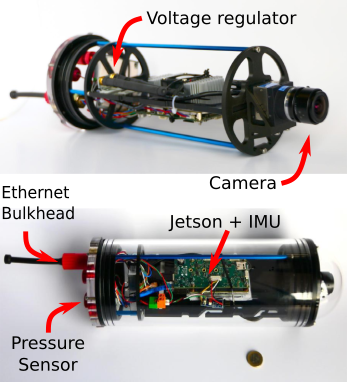}
\caption{The acquisition system integrating a monocular monochromatic camera, a pressure sensor and an IMU.}
\label{fig:caisson}
\end{figure}

\noindent The acquisition system that we designed to record the dataset can be seen in figure \ref{fig:caisson}.  It consists of a watertight enclosure containing a monochromatic camera, a pressure sensor and an Nvidia Tegra Jeston TX2 module mounted on Auvidea J120-IMU carrier board.  Details are given in table \ref{tab:sensors}.  The monochromatic camera is equipped with a wide-angle lens and records images with a resolution of 640x512 pixels at 20 Hz.  The camera is set behind a dome end cap in order to reduce distortion from the passing of light rays through different media.  The IMU provides linear acceleration and angular velocity measurements at 200 Hz, as well as compass measurements at 80 Hz and the pressure sensor between 5 and 10 Hz.  The computing module is running Ubuntu 16.04 and records synchronously the different sensors measurements thanks to the ROS middleware.  An advantage of our payload is that it is independent of any robotic architecture and can thus be embedded on any kind of Remotely Operated Vehicle (ROV) or Autonomous Underwater Vehicle (AUV).  This contrasts with classical underwater robots instrumentation load which are usually integrated in the design of the robots \cite{sparusII}.

\begin{table}
\tiny
\centering
\begin{tabular}{@{}clll@{}}
\toprule
\multicolumn{1}{l}{}              & Specifications                     & Values        & Description               \\ \midrule
\multirow{4}{*}{Vehicle}          & Size                               & 65x60x55 cm            &                           \\
                                  & Mass                               & 28 Kg               &                           \\
                                  & Thrusters                          & 8              & Blue Robotics T200                       
\\ \cmidrule(l){1-4}

\multirow{17}{*}{Payload} 		  & \textbf{Camera sensor}             &               & uEye - UI-1240SE          \\
                                  & Resolution                         & 640x512 px    &                           \\
                                  & Sensor                             & Monochromatic &                           \\
                                  & Frame per second                   & 20 Hz         &                           \\
                                  & \textbf{Lens}                      &               & Kowa LM4NCL C-Mount	   \\
                                  & Focal Length                       & 3.5mm         &                           \\
                                  & Field of View                      & 131$\deg$     &                           \\
                                  & \textbf{Inertial Measurement Unit} &               & MEMS - MPU-9250           \\
                                  & Gyroscope frequency                & 200 Hz        &                           \\
                                  & Accelerometer frequency            & 200 Hz        &                           \\
                                  & Compass                            & 80 Hz         &                           \\
                                  & \textbf{Pressure Sensor}           &               & MS5837 - 30BA             \\
                                  & Depth Range                        & 0 - 300m      &                           \\
                                  & Resolution                         & 0.2 mbar      &                           \\
                                  & Frequency                          & 5-10 Hz       &                           \\
                                  & \textbf{Embedded Computer}         &               &                           \\
                                  & Computing Unit                     &               & Nvidia - Tegra Jetson TX2 \\
                                  & Carrier board                      &               & Auvidea J120 - IMU                \\
                                  & \textbf{Housing}         &               &                           \\
 						  		  & Enclosure						   & 33.4 x 11.4 cm & 4" Blue Robotics Watertight Enclosure 
\\
						  		  & Enclosure End Cap				   & Dome & 4" Blue Robotics Dome End Cap 

\\ \cmidrule(l){1-4} 
\end{tabular}
\caption{Summary of the features of the ROV and the payload.}
\label{tab:sensors}
\vspace{-7.5mm}
\end{table}

\section{Dataset}
\label{sec:data}

\begin{figure}
\centering
\includegraphics[width=0.8\linewidth]{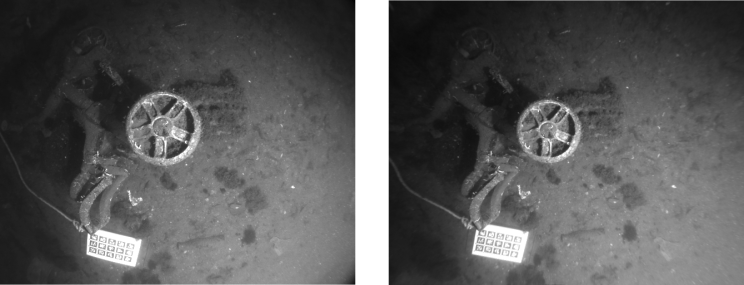}
\caption{Equidistant distortion effects and removal from Kalibr calibration.  Left: raw image.  Right: undistorted image.}
\label{fig:kalibr}
\vspace{-7.5mm}
\end{figure}

\noindent In order to record the dataset, the acquisition system is set to face downward on the ROV as shown in figure \ref{fig:rov_and_caisson}.  An aprilgrid is used both for the calibration of the system and as a marker for navigation.  In fact, in each sequence, the ROV starts from the aprilgrid (visible), navigates away from it (no-more visible), and finally comes back to it (visible again).  Intrinsic calibration of the camera is done in situ while calibration of the extrinsic parameters between the IMU and the camera is performed in-air in order to make fast motions, required to estimate these parameters.  These calibration steps are computed using  Kalibr \cite{Kalibr}.  As the camera is equipped with a wide-angle camera, we used the equidistant distortion model of Kalibr.  Results of the calibration can be seen figure \ref{fig:kalibr}.  

\noindent All the measurements have been recorded with the ROS middleware and they are hence all synchronized in ROS bags.  
In the online repository of the dataset, we provide both the data in a ROS bag format and as plain files.  For the plain files, the IMU, pressure and compass measurements are given in independent files with a timestamp linked to every measure.  The distorted and undistorted version of the images are also given as \textit{png} images and the timestamp of each image is written in a separate file.  

\begin{figure*}[!t]
\centering
\includegraphics[width=\linewidth]{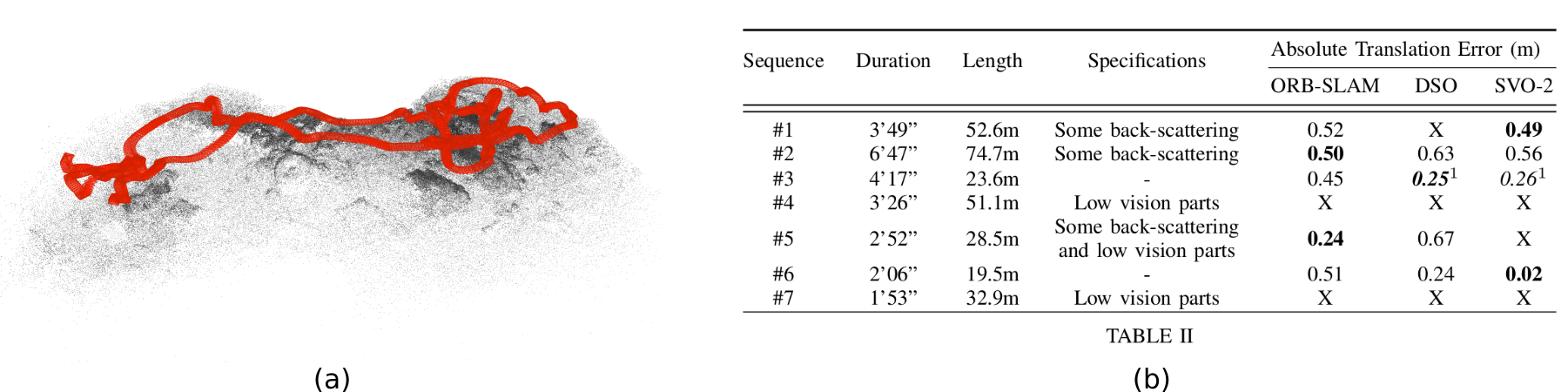}
\caption{(a) Sequence \#3 Colmap 3D reconstruction. (b) Evaluation of state-of-the-arts monocular VSLAM methods on the recorded dataset \textit{(1:  Results obtained starting the sequence 15s later otherwise the methods failed to initialize)}.}
\label{fig:results}
\end{figure*}

\noindent A total of seven sequences have been recorded using this setup.  The sequences were recorded in an harbor in collaboration with the DRASSM\footnote{DRASSM: French Department of Underwater Archaeological Research}.  The sequences expose different levels of difficulty with sometimes parts where vision becomes unusable.  Details about each sequence are given in Fig.\ref{fig:results}b.  Obtaining a ground-truth is very difficult in underwater environments and often requires the use of external infrastructures.  However, Structure-from-Motion methods are able to compute very accurate 3D reconstruction from sequence of images by means of extensive full batch Bundle Adjustment run offline.  In order to compute such a reconstruction we have used the state-of-the-art library Colmap \cite{Colmap}.  Setting very low the features detection threshold, Colmap has been able to produce very accurate reconstruction (Fig.\ref{fig:results}a).  As the poses of every image is computed by Colmap in order to produce the 3D reconstructions, we have extracted the estimated cameras' trajectories to use it as a ground-truth.  We further scaled these trajectories using depth measurements to get metric trajectories.

\section{Benchmark}
\label{sec:eval}

\noindent As a benchmark, we have run experiments using state-of-the-art monocular VO and VSLAM algorithms on each sequences.  We compare ORB-SLAM \cite{ORB-SLAM}, SVO-2 \cite{SVO-2} and DSO \cite{DSO}.
ORB-SLAM is a feature-based SLAM algorithm extracting ORB \cite{Orb} features for tracking and for loop closure detections.  It makes use of keyframes, connected to each others from their feature matches, and efficiently optimize the built graph through Bundle Adjustment.  
SVO-2 is a semi-direct sparse odometry method.  Direct methods do not rely on matched features as a mean of tracking and pose estimation but instead directly track photometric patches in the images.  SVO-2 is semi-direct in the sense that FAST features \cite{Fast} are detected in each new keyframe and then image-alignment is performed using photometric patches around the features to estimate the pose.
DSO is a pure direct odometry algorithm.  It performs pose estimation from the minimization of photometric errors across several images by tracking photometric patches of areas with high intensity gradients.  As the tracking is direct, the algorithm takes into account illumination changes in the energy minimization step, thus relaxing the constant brightness assumption of many direct methods.

\noindent The results for each method are given in Fig.\ref{fig:results}b.  As any monocular systems, trajectories are estimated up to scale.  In order to compute an absolute translation error for every sequence, the monocular trajectories are first scaled by applying the similarity transformation that best fit the ground-truth.
ORB-SLAM is the most stable algorithm on this dataset as it manages to run on five out of the seven sequences while DSO and SVO-2 run only on four sequences each.  Interestingly, ORB-SLAM does not manage to detect any loop closure in the sequences, thus asking the question of the relevance of its loop detection mechanism for underwater environments.

\noindent These results highlight the potential of vision based localization methods for underwater environments.  While not all the sequences can be processed, the result are promising.  Both the direct and feature-based paradigms seem to work but their seem to be a compromise between stability and accuracy.  A feature-based SLAM algorithm combined to a direct tracking of the features could be a good balance for underwater localization.

\section{Conclusion}

In this paper, we have presented a new underwater dataset focusing on vision-based localization methods and including IMU and pressure measurements.  We made publicly available this dataset to the benefit of the community.  The designed acquisition system as well as the recorded sequences were then presented in detail.  Finally, we gave the results in terms of localization drift for VO and VSLAM state-of-the-art methods on each sequence.  These results gave a short overview of what can be achieved using pure visual information and could be used as a benchmark on this dataset.  There is room for improvement as the evaluated methods were originally developed for terrestrial and aerial applications.  Therefore, methods specifically dedicated to underwater visual degradations and the integration of the IMU and pressure measurements are expected to give very interesting results.  In future work, we will investigate the use of such SLAM methods.

\tiny
\bibliographystyle{IEEEtran}
\bibliography{biblio}

\end{document}